\pdfoutput=1
\documentclass[11pt]{article}
\usepackage{acl}
\usepackage{times}
\usepackage{url}
\usepackage{amsmath}
\usepackage{times}
\usepackage{graphicx}
\usepackage{xcolor}
\usepackage{latexsym}
\usepackage[capitalize,noabbrev]{cleveref}
\usepackage{microtype}
\usepackage[T1]{fontenc}
\usepackage[utf8]{inputenc}
\usepackage{tabularx}
\usepackage{booktabs}
\usepackage{multirow}
\usepackage{xspace}
\usepackage{floatrow}
\usepackage{enumitem}
\setlist[enumerate]{itemsep=0mm}

\graphicspath{{img/}}

\setkeys{Gin}{width=.94\textwidth}

\makeatletter
\g@addto@macro\@floatboxreset\centering
\makeatother

\usepackage{xspace}

\makeatletter
\DeclareRobustCommand\onedot{\futurelet\@let@token\@onedot}
\def\@onedot{\ifx\@let@token.\else.\null\fi\xspace}

\def\eg{e.g\onedot} 
\def\ie{i.e\onedot} 
 
 \def\vs{vs\onedot}

\makeatother

\title{Scalable Performance Analysis for Vision-Language Models}

\author{Santiago Castro\(^*\) \quad Oana Ignat\(^*\) \quad Rada Mihalcea \\
University of Michigan -- Ann Arbor, USA \\
\texttt{\{sacastro,oignat,mihalcea\}@umich.edu}}

\begin{document}

\maketitle

\renewcommand*{\thefootnote}{\fnsymbol{footnote}}
\footnotetext[1]{Equal contribution.}
\renewcommand*{\thefootnote}{\arabic{footnote}}
\setcounter{footnote}{0}

\begin{abstract}
Joint vision-language models have shown great performance over a diverse set of tasks. However, little is known about their limitations, as the high dimensional space learned by these models makes it difficult to identify semantic errors. Recent work has addressed this problem by designing highly controlled probing task benchmarks. Our paper introduces a more scalable solution that relies on already annotated benchmarks. Our method consists of extracting a large set of diverse features from a vision-language benchmark and measuring their correlation with the output of the target model. We confirm previous findings that CLIP behaves like a bag of words model and performs better with nouns and verbs; we also uncover novel insights such as CLIP getting confused by concrete words. Our framework is available at \url{https://github.com/MichiganNLP/Scalable-VLM-Probing} and can be used with other multimodal models and benchmarks.
\end{abstract}

\section{Introduction}

Recent years have witnessed an explosion of vision-language models~\cite{Lu2019ViLBERTPT,Li2019VisualBERTAS,Zhang2021VinVLRV,Radford2021LearningTV,Singh2021FLAVAAF}.
These models have shown great performance in a variety of tasks, such as image/video classification and text-image/video retrieval  \cite{Radford2021LearningTV,clip4clip}, even without leveraging task-specific or in-domain training. In addition, these models have shown to be practical when leveraged as underlying models for text-to-image generation such as DALL-E~2~\cite{ramesh2022hierarchical} and image captioning such as ClipCap~\cite{mokady2021clipcap}.

Little is however known about the limitations of these models.
Recent work, such as Winoground~\cite{Thrush2022WinogroundPV}, SVO-Probes~\cite{Hendricks2021ProbingIT}, or VALSE~\cite{Parcalabescu2022VALSEAT}, have designed benchmark probing tasks by annotating data to follow specific properties (\ie{}, object color, location, size, swapping word order, replacing words). 
This line of research led to valuable insights into the limitations of current state-of-the-art multi-modal models such as CLIP~\cite{Radford2021LearningTV} and ViLBERT~\cite{Lu2019ViLBERTPT}.

\begin{figure}
\includegraphics{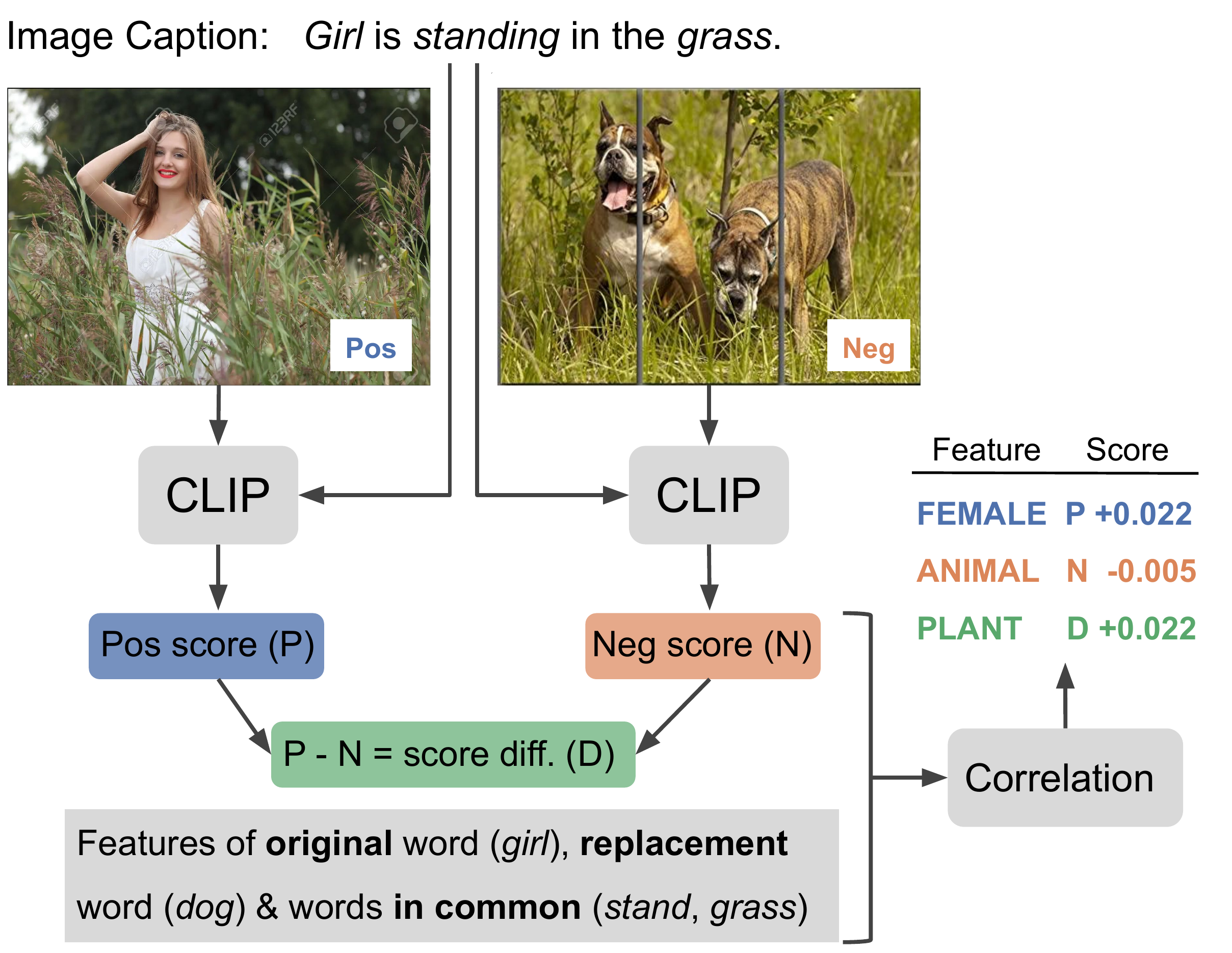}
\caption{We propose a simple framework to analyze CLIP performance on SVO-Probes data. We test CLIP on the benchmark, extract a diverse set of semantic features from the data, and measure the correlation between each feature and the CLIP score (\(P\), \(N\), or \(D\)). Features with positive correlation (\eg{}, \textit{Female, Plant}) impact positively the model performance, while features with negative correlation (\eg{}, \textit{Animal}) impact negatively the model performance.}
\label{fig:overview}
\end{figure}

An important limitation of current work is the reliance on time-consuming data annotation procedures, making it unscalable and limited in scope. As a complementary solution, we propose a method to probe vision-language models by relying on existing data, without requiring extra annotations.
The method consists of extracting a large set of candidate features from a vision-language benchmark and testing their correlation with respect to the output of the target models on the given benchmark.

By applying our method on CLIP~\cite{Radford2021LearningTV}, a widely used state-of-the-art multi-modal model, using the SVO-Probes~\cite{Hendricks2021ProbingIT} benchmark, we confirm the findings of \citet{Thrush2022WinogroundPV} of CLIP behaving like a bag of words model and that of \citet{Parcalabescu2022VALSEAT} of CLIP performing better with nouns and verbs.
We also find that CLIP gets confused by concrete words and that it surprisingly improves in performance for more ambiguous words while noting little change from the word frequencies.
To the best of our knowledge, we are the first to conduct an in-depth analysis of how language semantic properties influence CLIP's performance.

We summarize our contributions as follows. 
First, we propose a scalable way of measuring the limitations of vision-language models. Second, we test our method using a state-of-the-art vision-language model (CLIP) and a popular benchmark (SVO-Probes), validate known challenges, and uncover new ones. Third, our work opens up avenues for future models to focus on solving the newly discovered challenges.

\section{Related Work}

Recently, an increasing number of benchmarks have been created for the evaluation of vision-language model abilities to perform various multi-modal tasks.

\citet{Hendricks2021ProbingIT} evaluate state-of-the-art vision-language models 
by building SVO-Probes, a probing benchmark focused on verb understanding. They show that image–language transformers fail to distinguish fine-grained differences between images and find that they are worse at verb understanding compared to subjects or objects. In our work, we continue their proposed future work direction by analyzing model performance on fine-grained verb categories.

Other work focuses on testing more precise capabilities of vision-language models using other probing techniques. In VALSE, \citet{Parcalabescu2022VALSEAT} demonstrate that vision-language models have difficulty in counting objects and in correctly classifying spatial relations between objects. 
\citet{Salin2022AreVT,Zhao2022VLCheckListEP} show that, although state-of-the-art vision-language models can grasp color, they do not fully understand more difficult concepts such as object size and position in the image.

In Winoground, \citet{Thrush2022WinogroundPV} designed adversarial examples that require differentiating between a similar image and text, where the text pairs only differ in their word order. Their results show that state-of-the-art vision-language models lack compositional reasoning abilities.
Several other works build benchmarks on probing vision-language on compositional reasoning~\cite{Akula2020WordsAE,Ma2022CREPECV,Liu2022VisualSR,Park2022ExposingTL,Yuksekgonul2022WhenAW} find that they behave like a bag-of-words model -- \ie{}, have poor relational understanding and a severe lack of word order sensitivity. 

In contrast, our work focuses not on creating new probing tasks for vision-language models, but on using current benchmarks to learn additional, more fine-grained features that can be discovered using simple correlation methods. 
To the best of our knowledge, we are the first to analyze the performance of CLIP on a diverse set of semantic features 
and use correlation methods to draw insights about what concepts are challenging for the model.

\section{Methodology to Probe CLIP}

Given a benchmark, we measure how a vision-language model performs on a variety of semantic concepts. Our aim is to quantify which concepts are the most and the least challenging for the model. 
Our setting is illustrated in \cref{fig:overview}, and can be described in three main steps.

First, we use CLIP~\cite{Radford2021LearningTV} to compute scores for instances from the SVO-Probes~\cite{Hendricks2021ProbingIT} dataset and obtain two corresponding alignment scores for each sentence and its corresponding \textit{positive} and \textit{negative} image. Next, we extract and process a diverse set of semantic features from SVO-Probes. Finally, we compute the correlation coefficients between each feature and the CLIP score. The features with the highest coefficients will represent concepts that CLIP performs well on, while features with the lowest coefficients will represent challenging concepts for CLIP.

\subsection{Dataset}

We choose the SVO-Probes~\cite{Hendricks2021ProbingIT} dataset due to its design and large scale size (421 verbs and over 48,000 image-sentence pairs).
SVO-Probes was designed for probing image-text models for their understanding of \textbf{s}ubject, \textbf{v}erb, \textbf{o}bject triplets.
Each instance from the dataset consists of a text caption, a \textit{positive} image that matches the caption, and a controlled (adversarial) \textit{negative} image that shares two out of three aspects (subject, verb, and object) from the sentence but does not match the other one, as shown in \cref{fig:overview}. %
These controlled examples enable one to probe models for their understanding of verbs as well as subjects and objects.
The instances also include information about the negative image, such as a (hidden) associated negative caption which we leverage in this paper.

We propose to use this dataset to evaluate the CLIP~\cite{Radford2021LearningTV} model. 
We choose to test CLIP, as opposed to other language-vision models, due to its widely-spread use and impressive zero-shot performance on a variety of vision-language tasks (\eg{}, text-to-image retrieval, image question answering, human action segmentation, image-sentence alignment -- \citealt{Cafagna2021WhatVM}). 
Furthermore, \citet{Hendricks2021ProbingIT} test only ViLBERT-based~\cite{Lu2019ViLBERTPT} models, which are known to perform worse than CLIP~\cite{Cafagna2021WhatVM}.

\subsection{Model Output}%
\label{sec:model_output}

As depicted in \cref{fig:overview}, we obtain three CLIP scores for each pair of \textit{positive} and \textit{negative} images: a \textit{positive} score (\(P\)), computed between the caption and the \textit{positive} image; a \textit{negative} score (\(N\)), computed between the caption and the \textit{negative} image; and the \textit{difference} between these scores (\(D = P - N\)).

Because the text and the positive image are aligned, \(P\) represents an absolute alignment score.
In the case of the text and the negative image, even though the negative image is similar in some ways to the text (because of how SVO-Probes was designed), they do not correspond to each other. Thus, \(N\) represents an absolute misalignment score.
\(D\) represents a relative alignment score.
Ideally, CLIP should have a high \(P\) score and a low \(N\) score, and a high difference between them (a high \(D\)).
We propose to pay special attention to \(D\) given that CLIP is generally used in relative comparisons, such as when using it for classification (choosing the class text that maximizes the alignment score, given an image) or when using it for retrieval (finding the text/image that maximizes the alignment score given an image/text).

\subsection{Feature Extraction}%
\label{sec:features}

For each given sentence and corresponding image in the benchmark, we extract features from the words marked in the SVO-Probes benchmark (\ie{}, subject, verb, and object).

If the corresponding image is \textit{positive}, all the extracted features are from words \textit{in common}, \ie{}, that appear both in the image and the text. Otherwise, if the corresponding image is \textit{negative},
in addition to words \textit{in common}, we also 
extract features from words present in the sentence and not in the image (\textit{original} word) and words present in the image but not in the text (\textit{replacement} word).
As an example, in \cref{fig:overview} the words \textit{in common} are ``sit'' and ``grass'', the \textit{original} word is ``girl'' and the \textit{replacement} word is ``dogs''.
The \textit{original} and \textit{replacement} words represent what is different between the image and the text, while the words \textit{in common}, as the name suggests, represent what is common between the image and the text. 

We extract the following \textbf{semantic} textual features: \citet{Levin1993EnglishVC} verb classes,  LIWC psycholinguistic markers~\cite{Pennebaker2007LinguisticIA, Pennebaker2015TheDA},
General Inquirer~\cite{Stone1967TheGI} semantic classes, WordNet hypernyms~\cite{Miller1995WordNetAL}, word presence, semantic similarity, ambiguity, frequency, sentence length, and concreteness~\cite{brysbaert2014concreteness}.

\paragraph{Levin verb classes.}
\citet{Levin1993EnglishVC} groups verbs according to their semantic content and also according to their participation in argument alternations.

Levin's semantic content-based taxonomy provides a classification of 3,024 verbs into 48 broad classes and 192 fine-grained classes.\footnote{\url{https://websites.umich.edu/~jlawler/levin.verbs}}
A verb can belong to one or more classes.
Some examples of verb classes are: (1) broad \textit{change of state} (\eg{}, clean, divide, soak), \textit{manner of motion} (\eg{}, climb, drop, run) or \textit{social interaction} (\eg{}, marry, meet, hug); 
(2) fine-grained: \textit{``roll'' verbs} (\eg{}, bounce, coil, drift), \textit{``run'' verbs} (\eg{}, amble, bolt, race) or \textit{``hug'' verbs} (\eg{}, cover, encircle, touch)

\paragraph{LIWC psycholinguistic markers.}
Linguistic Inquiry and Word Count (LIWC)~\cite{Pennebaker2007LinguisticIA,Pennebaker2015TheDA} is a widely used word-counting software that includes dictionaries of English words related to human cognitive processes. 
Specifically, we use the LIWC2015 dictionary, which contains 6,400 words and word stems. Each word or word stem defines one or more categories: \eg{}, the word ``mother'' is assigned the categories: \textit{female, family, social}.

\paragraph{General Inquirer classes.}
General Inquirer~\cite{Stone1967TheGI} is a resource for automatic content analysis. More specifically, it categorizes words into emotional and cognitive states, as well as into diverse semantic categories outlined in the Lasswell dictionary~\cite[pg. 46--53]{Namenwirth1986DynamicsOC}.

\paragraph{WordNet classes.}
WordNet~\cite{Miller1995WordNetAL} is a large lexical database of English words that are grouped into sets of cognitive synonyms, known as synsets. The synsets are interlinked by semantic and lexical relations. The most frequent relation among synsets is the super-subordinate relation, also called \textit{hyperonymy}. It links more general synsets to specific ones: \eg{}, ``building'' is a \textit{hypernym} of ``house'' and ``school''. 
For each given word, we collect all the hypernyms of the most common word synset.

\paragraph{Word presence.} For each given word, we use a marker to indicate if the word is present or not in the sentence. Note that studying the effect of specific words does not imply that they have no dependencies with other words. Their role may change depending on the context; however, we study them in aggregate.

\paragraph{Sentence length.} We measure the length of each sentence as the number of words in the sentence.

\paragraph{Semantic similarity.}
In the case of \textit{negative} images, we compute the cosine similarity score between the \textit{original} words and the corresponding \textit{replacement} words. The word representations are computed using Sentence-Transformers~\cite{reimers-2019-sentence-bert}, with the model \texttt{all-MiniLM-L6-v2}, which is based on MiniLM~\cite{minilm}.

\paragraph{Concreteness score.}
For measuring the concreteness of words, we use a dataset of words with associated concreteness scores from \citet{brysbaert2014concreteness}. Each word is labeled by a human annotator with a value between 1 (very abstract) and 5 (very concrete). Abstract words (\eg{}, ``beauty", ``sadness'') denote ideas, feelings, or other intangible concepts while concrete words (\eg{}, ``table", ``write'') refer to objects and actions. 

\paragraph{Ambiguity.}
We measure the ambiguity of a given word by counting the number of synsets in WordNet~\cite{Miller1995WordNetAL}.

\paragraph{Frequency.}
We measure the word frequency in a subset (\(\sim\)13M image captions) of LAION~\cite{Schuhmann2021LAION400MOD}, a dataset representative of CLIP's training data. 

\subsection{Feature Representation}

The \textbf{binary} features, \ie{}, Levin, LIWC, General Inquirer, WordNet classes, and word presence, are represented as binary vectors, while the \textbf{numerical} features \ie{}, sentence length, concreteness, similarity, ambiguity, and frequency are standardized.
All the features are then concatenated together.

\subsection{Feature Selection}%
\label{sec:feature-selection}
We measure the degree of correlation between each feature and the model performance.
For each of the \textbf{binary} features, we compute a two-sample two-tailed t-test \citep{student1908probable} along with the model output score.
This test evaluates if the means of the populations coming from each feature value (true or false) are significantly different.
If so, we compute the difference of means as a reference value.
In the case of \textbf{numerical} features, we compute the Pearson's correlation coefficient \citep{benesty2009pearson} between each feature and the model performance score. 

Next, we employ a one-sample, two-tailed t-test to determine if the coefficient is significantly different from zero, \ie{}, if there is any correlation according to this metric.
We chose a p-value threshold of 0.05 (a confidence level of 95\%) to filter out the features.\footnote{See the obtained scores and p-values in the web page from this paper: \url{https://github.com/MichiganNLP/Scalable-VLM-Probing}.}

\subsection{Experimental Details}

We use an OpenAI pre-trained CLIP~\citep{Radford2021LearningTV} ViT-L/14~\citep{vit} model.

\nocite{Paszke_PyTorch_An_Imperative_2019,harris2020array,The_pandas_development_team_pandas-dev_pandas_Pandas,Lhoest_Datasets_A_Community_2021,casper_da_costa_luis_2023_7697295,seabold2010statsmodels,Bird_Natural_Language_Processing_2009,Waskom2021,Honnibal_spaCy_Industrial-strength_Natural_2020,hunter2007matplotlib,Wolf_Transformers_State-of-the-Art_Natural_2020,virtanen2020scipy,scikit-learn}

\section{Results}

Our main observations and takeaways from this evaluation are the following:

\paragraph{(1) CLIP behaves like a bag-of-words model.}

\begin{figure}
\includegraphics{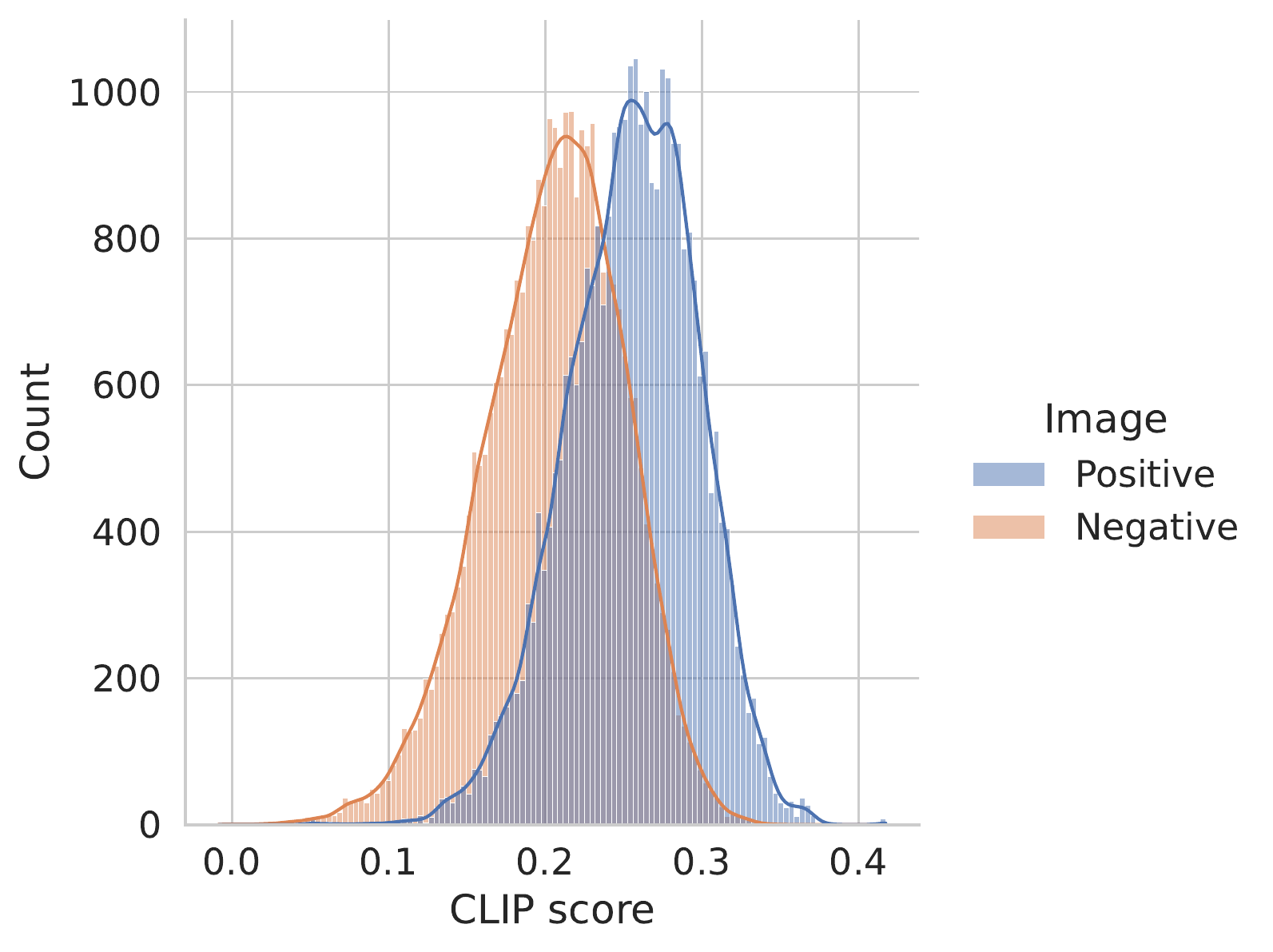}
\caption{Histogram plot of the distribution of CLIP scores between the text with the positive image, and the text with the negative image. A kernel density estimation curve is included to aid this visualization.}
\label{fig:score_dist}
\end{figure}

As shown in \cref{fig:score_dist}, the distributions of \(P\) and \(N\) highly overlap.
This is explained partly by the negative image being adversarial; it contains elements in common with the text.
This finding is coherent with that of \citet{Thrush2022WinogroundPV}, that CLIP performs like a bag-of-words model.

This finding is also supported by the fact that many features from words \textit{in common} contribute to increasing both the positive (\(P\)) and the negative scores (\(N\)): \eg{}, hypernym\_food.n.02 increases \(P\) by 0.042 and \(N\) by \(0.050\); LIWC ``money'' increases \(P\) by 0.036, and \(N\) by \(0.032\).
As described in \cref{sec:feature-selection}, we measure the importance of each feature as the difference of means between the CLIP scores when the feature is present and when is not.
We observed that many of the features for the words \textit{in common} appeared to influence similarly both \(P\) and \(N\), confirming this hypothesis.

\paragraph{(2) CLIP performs better with nouns than with verbs.}

When computing the number of times CLIP assigns a higher score to the similarity between the text and the \textit{positive} image as compared to the similarity between the text and the \textit{negative} image, the verbs obtain 81.45\% accuracy while the subjects get 86.87\% and the objects 88.78\%.
The number obtained for verbs is relatively close to that of a similar setting experimented by the VALSE benchmark~\citep{Parcalabescu2022VALSEAT}, in which they reported 75.6\% accuracy (also considering that we could not determine which pre-trained CLIP variant the authors evaluated). At the same time, the noun (objects and subjects) replacement numbers are consistent with those reported by the same authors (88.8\%), obtained from FOIL it!~\citep{foil_it}.

\paragraph{(3) CLIP gets confused by concrete words.}

\begin{figure}
\includegraphics{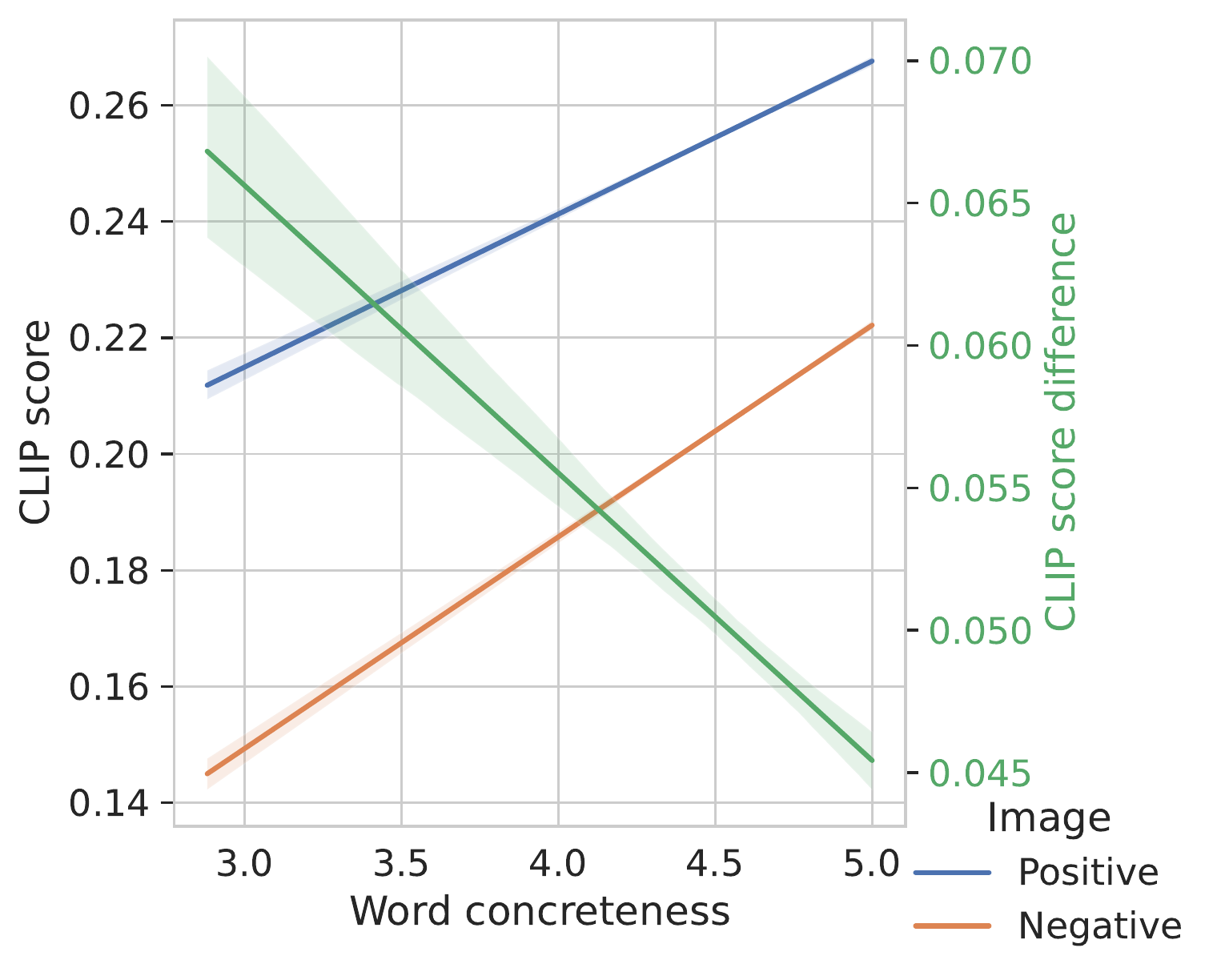}
\caption{Linear regression plot of the average concreteness for the words in the sentence that are common to both images \vs{} the CLIP score. The shadowed areas are 95\%-confidence intervals for the expected value.}
\label{fig:concreteness}
\end{figure}

\Cref{fig:concreteness} shows both the \textit{positive} and \textit{negative} CLIP scores improve the more concrete a word is (words from the caption represented in both the positive and the negative images). As seen in this figure, however, the \textit{negative} score increases faster.
This implies that, in an image classification or image-to-text retrieval setting, CLIP will more likely consider an incorrect text as correct if it has more concrete words than the actual correct text.

\paragraph{(4) CLIP prefers average-length sentences.}

\begin{figure}
\includegraphics{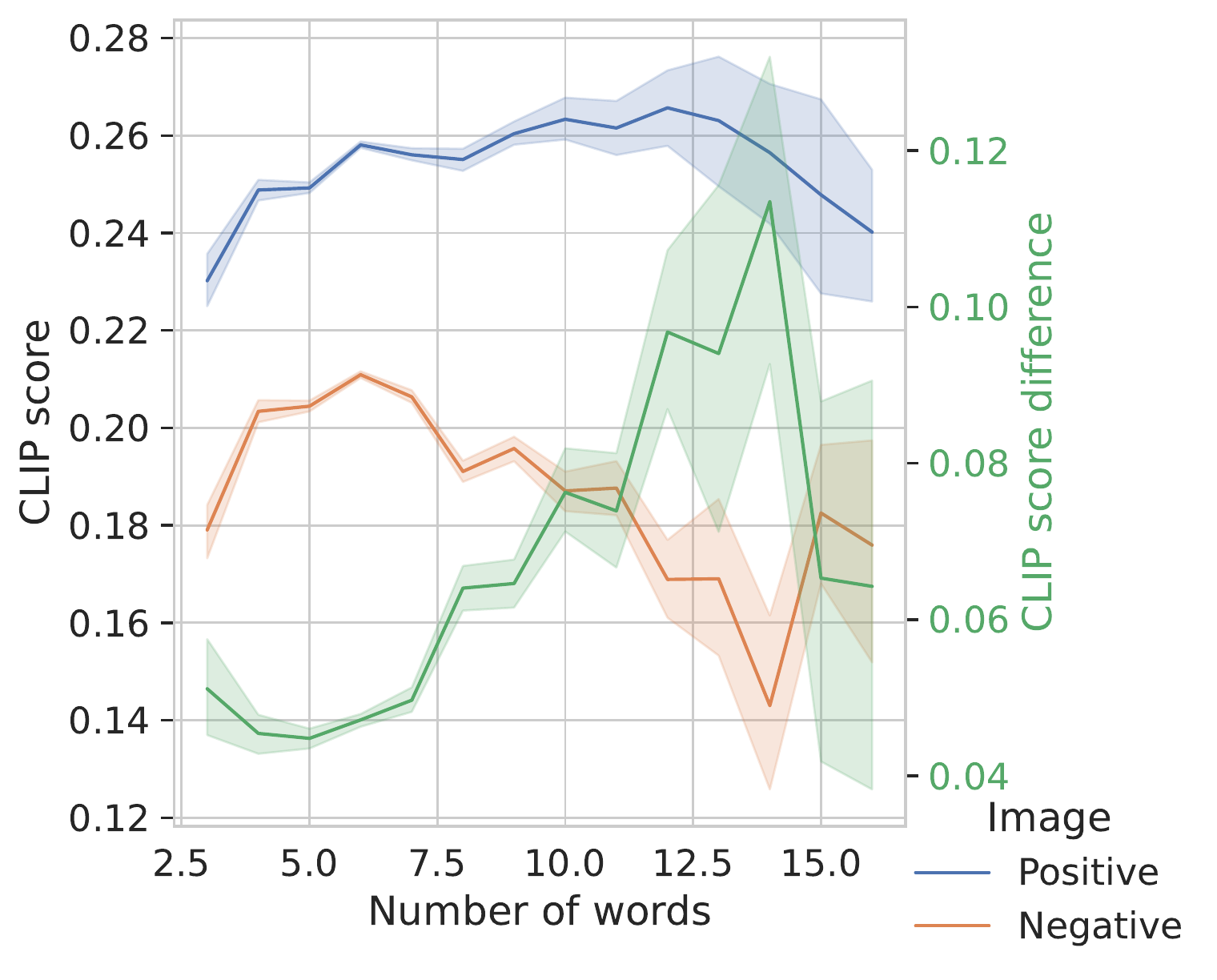}
\caption{Line plot of the number of words in the caption sentence \vs{} the CLIP score. The shadowed areas are 95\%-confidence intervals for the expected value.}
\label{fig:word_count}
\end{figure}

We present in \cref{fig:word_count} how the score is affected by the caption sentence word length.
CLIP presents a low performance when the sentences are very short (around 3 words long), improving when the sentences are longer since the difference between the \textit{positive} and \textit{negative} scores (\(D\)) gets larger with the sentence length.

\begin{figure}
\includegraphics{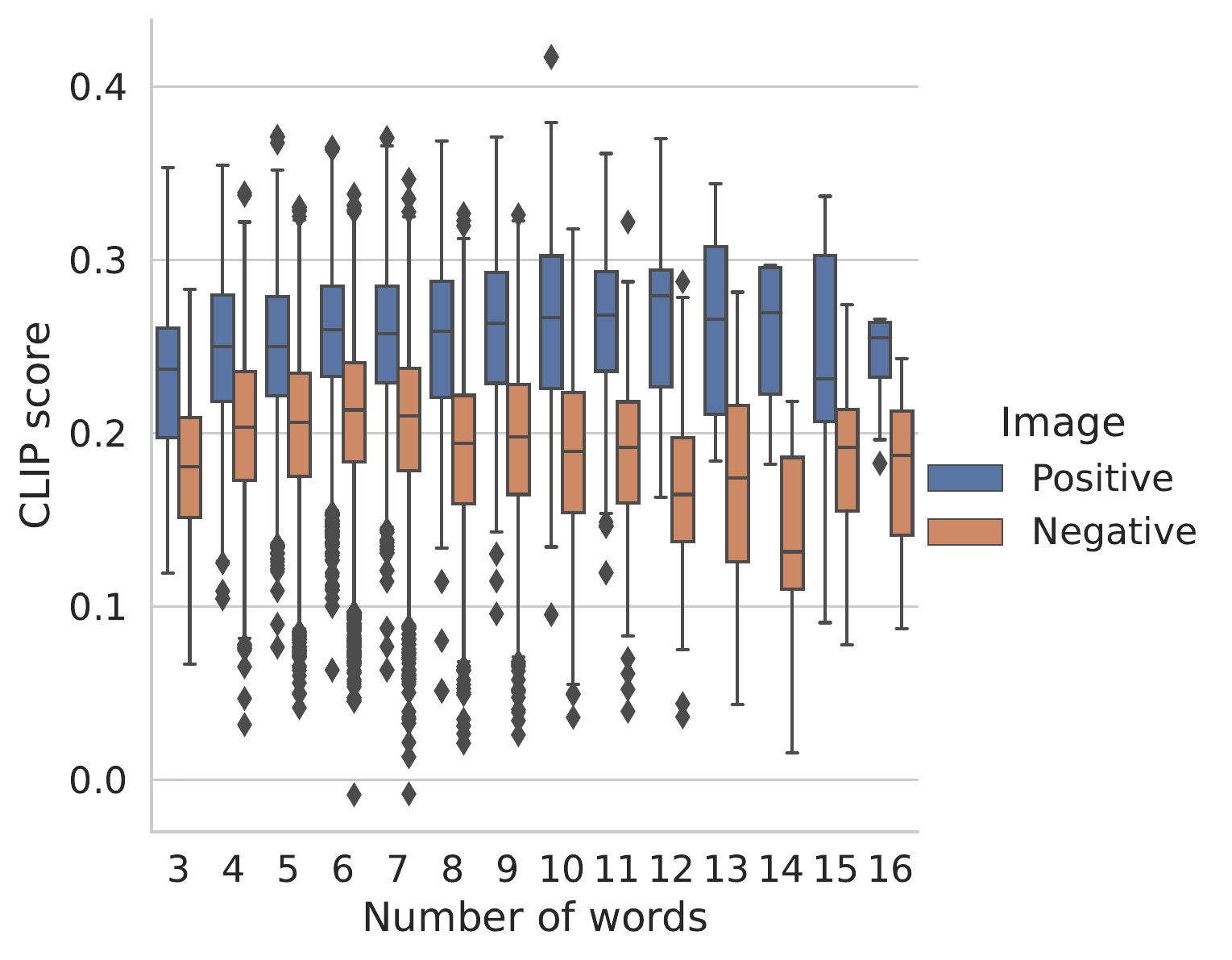}
\caption{Box plot for the number of words in the caption sentence \vs{} the CLIP score. Unlike \cref{fig:word_count} that shows the expected values, this plot shows the distributions.}
\label{fig:word_count_boxes}
\end{figure}

\Cref{fig:word_count_boxes} shows how the CLIP scores are distributed for the different number of words, showing for example that there is a great overlap between the similarity scores between texts of length 6 and a \textit{negative} image, and the similarity scores between texts of length 3 and a \textit{positive} image.
This implies CLIP is more likely to select the wrong text when comparing an image with a short correct text and one with long incorrect text.

\paragraph{(5) CLIP is affected by word frequency.}

\begin{figure}
\includegraphics{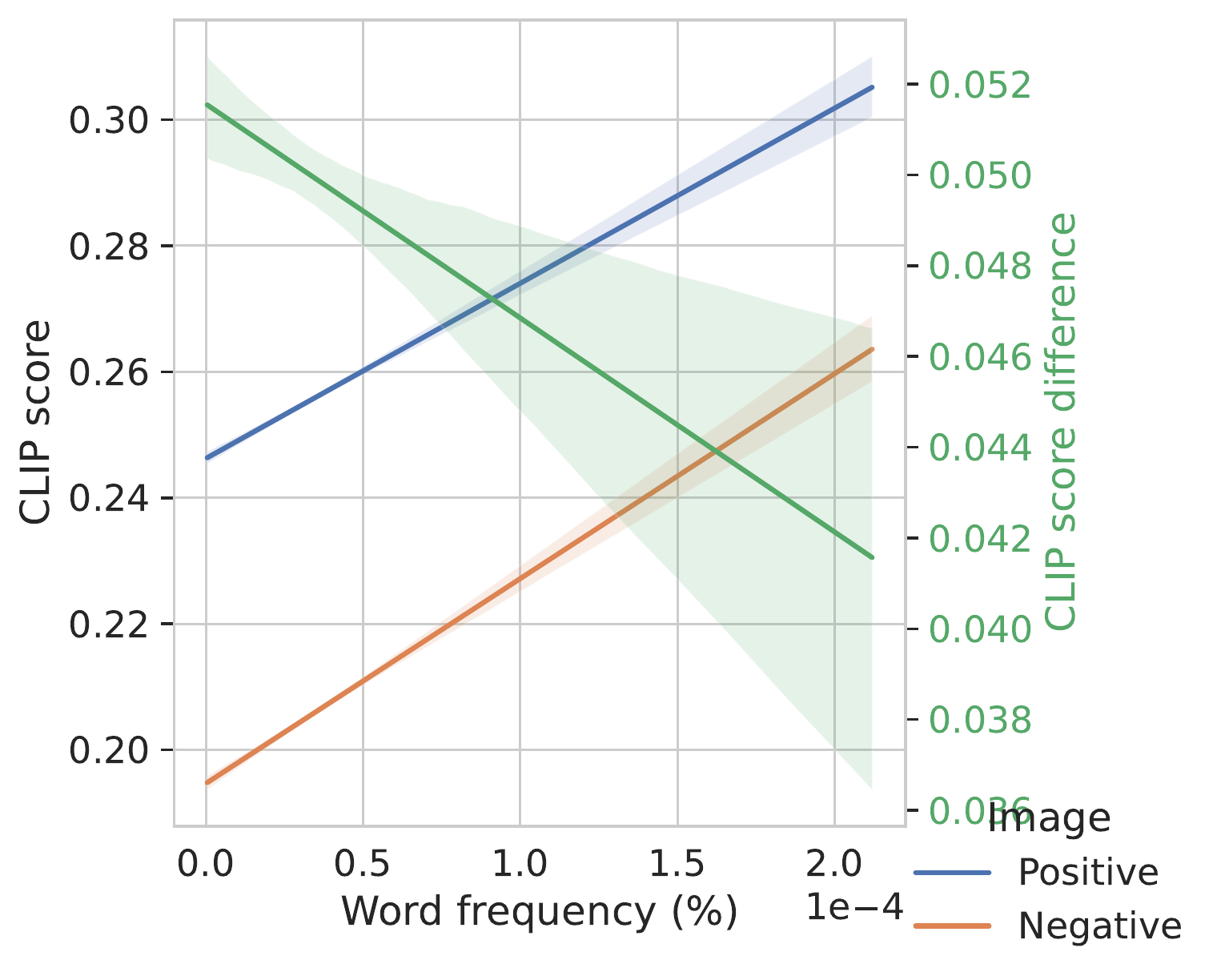}
\caption{Linear regression plot of the average frequency for the words in the sentence that are common to both images \vs{} the CLIP score. The shadowed areas are 95\%-confidence intervals for the expected value.}
\label{fig:freq}
\end{figure}

\Cref{fig:freq} studies the frequency effect on the score for the words that represent concepts that appear in both the \textit{positive} and \textit{negative} images. The more frequent a word is, the higher the CLIP score. Still, the difference in scores is barely affected.

\paragraph{(6) The score improves for more ambiguous words.}

\begin{figure}
\includegraphics{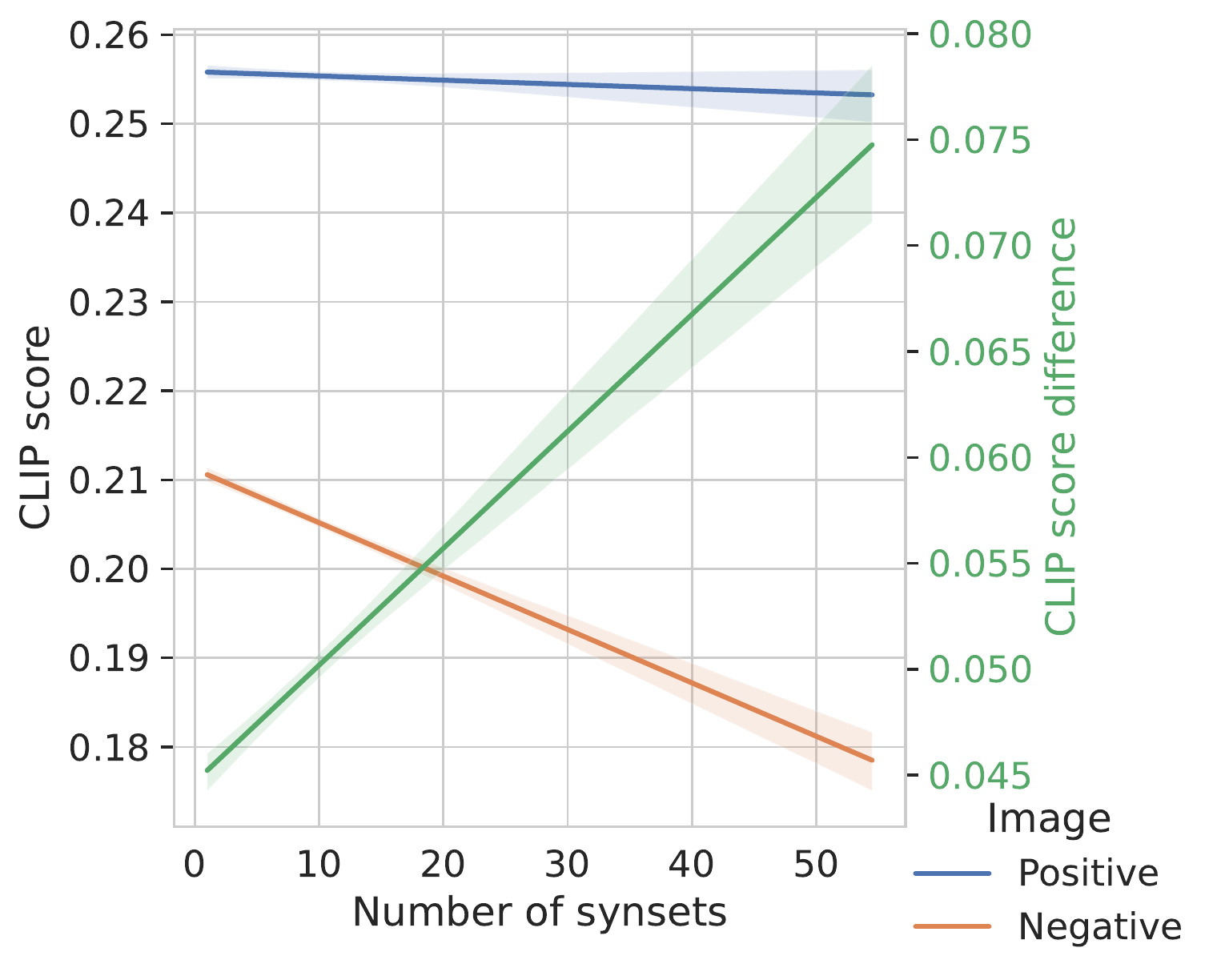}
\caption{Linear regression plot of the average synset count for the words in the sentence that are common to both images \vs{} the CLIP score. The shadowed areas are 95\%-confidence intervals for the expected value.}
\label{fig:synset_count}
\end{figure}

Surprisingly, there is a larger gap in the score difference (\(D\)) when the words have more meanings associated with them (for the words that represent concepts in both the \textit{positive} and \textit{negative images)}, as shown in \cref{fig:synset_count}. The positive score seems to remain almost constant while the negative score drops, widening the difference. The word frequency seems not to be a confounding factor based on (5).

\paragraph{(7) Similar situations confuse CLIP.}

\begin{figure}
\includegraphics{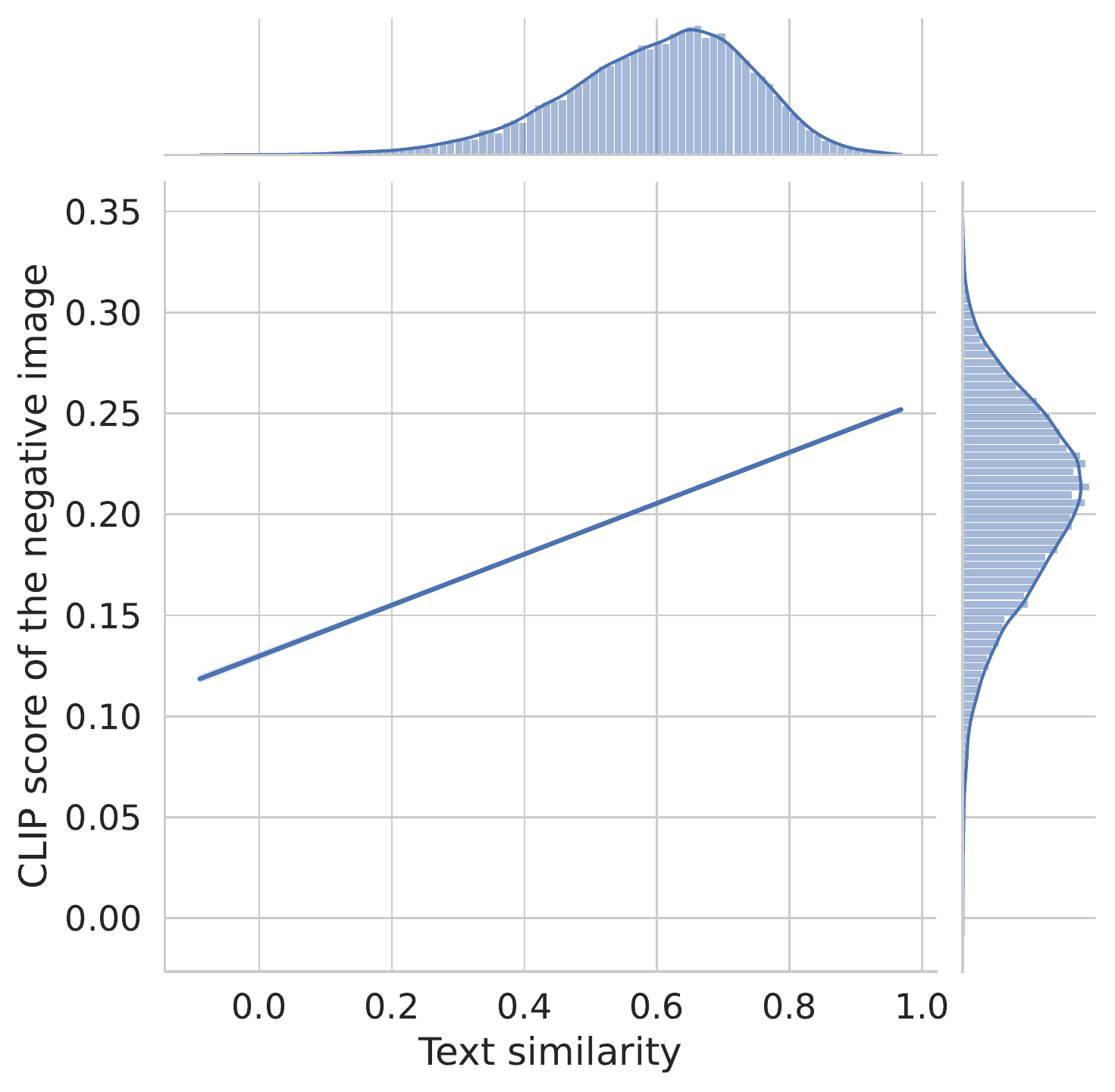}
\caption{Linear regression plot of the similarity between the text caption and the negative image text caption \vs{} the CLIP score for the negative image. The shadowed areas are 95\%-confidence intervals for the expected value. The unimodal distributions are also shown.}
\label{fig:text_similarity}
\end{figure}

Unsurprisingly, the higher the similarity between the caption and the negative image caption, the higher the \textit{negative} CLIP score, as depicted by \cref{fig:text_similarity}.

\begin{figure}
\includegraphics{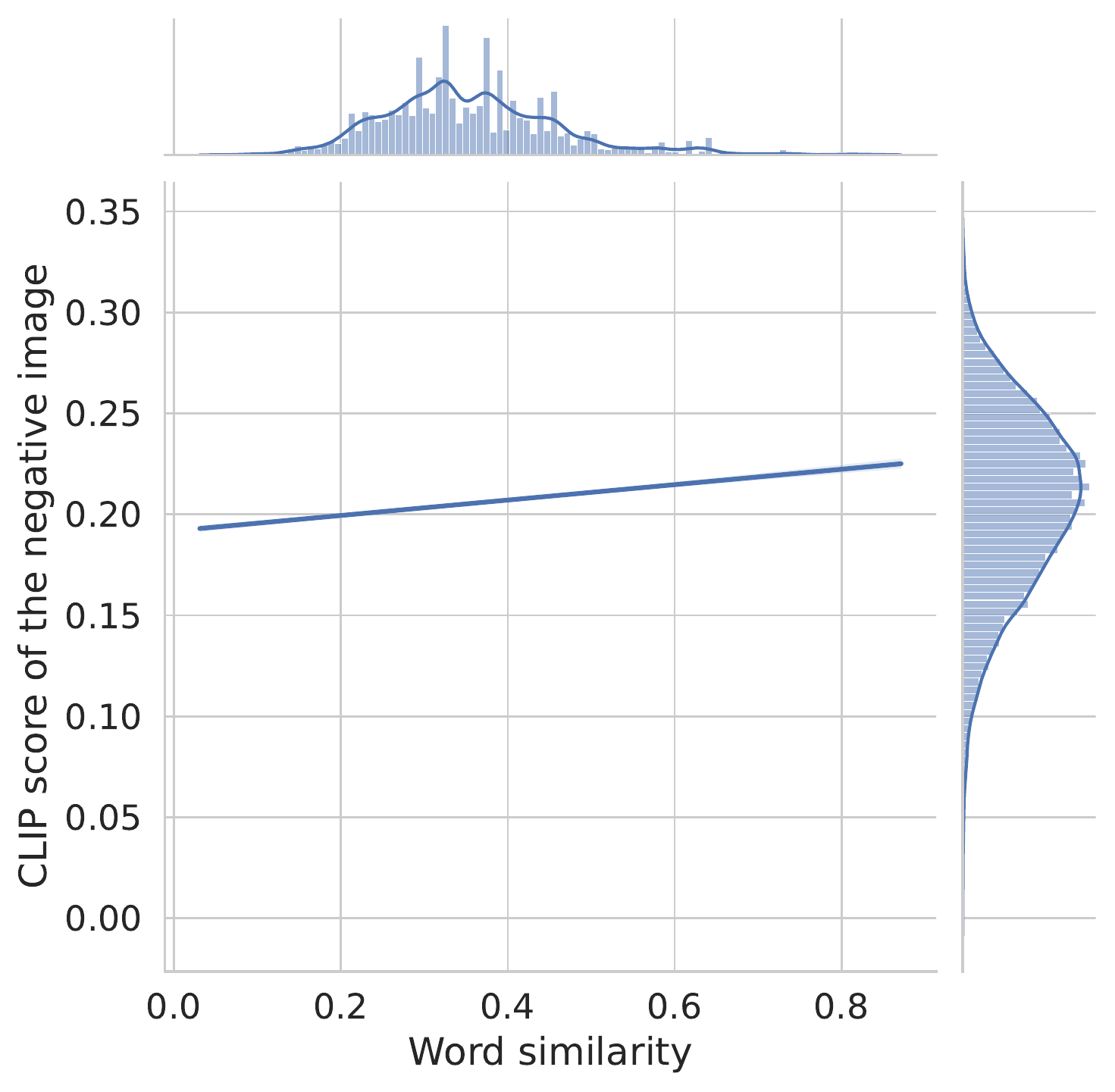}
\caption{Linear regression plot of the similarity between the originally replaced word from the text caption and new word from the negative image text caption \vs{} the CLIP score for the negative image. The shadowed areas are 95\%-confidence intervals for the expected value. The unimodal distributions are also shown.}
\label{fig:word_similarity}
\end{figure}

We also studied the influence of the similarity between the \textit{original} word (from the caption) and the \textit{replacement} word (from the text associated with the negative image) in \cref{fig:word_similarity}.
The effect of the word change seems to be smaller than that of the whole sentence change.

\renewcommand{\arraystretch}{1.2}
\begin{table*}
\footnotesize
\begin{tabular}{p{0.12\linewidth} p{0.42\linewidth} r p{0.27\linewidth}}
\toprule
Topic & Feature & Mean diff\onedot{} & Example Words \\
\midrule
\midrule
\multicolumn{4}{c}{\sc CLIP performs better on} \\
\midrule
\midrule
\multirow{2}{1em}{Natural Phenomenon} & Hypernym physical\_phenomenon.n.01 (original) & 0.038 & snow, fog, rain, mist \\ 
& Hypernym physical\_phenomenon.n.01 (replacement) & 0.022 & snow, rain, cloud, fog, mist \\ 
\hline
\multirow{2}{1em}{Waterfront Infrastructure} & Hypernym platform.n.01 (original) & 0.038 & pier, deck, podium \\
& Hypernym horizontal\_surface.n.01 (original) & 0.032 & pier, pavement,  quay\\
\hline
\multirow{5}{1em}{Landscapes} & Hypernym community.n.06 (original) & 0.038 & meadow, desert, grassland \\
& Hypernym natural\_elevation.n.01 (original) & 0.035 & dune, sandbar, reef \\
& Hypernym geological\_formation.n.01 (original) & 0.027 & beach, shore, cliff\\
& Hypernym plant.n.02 (original) & 0.025 & grass, tree, flower \\
& Hypernym natural\_elevation.n.01 (replacement) & 0.020 & mountain, hill \\
\hline
\multirow{4}{1em}{Grooming}& Presence of word ``wash'' (original) & 0.035 & wash\\
 & Levin ``floss verbs'' (original) & 0.030 & wash, brush, shave\\
 & Levin ``wipe verbs''(original) & 0.022 & wear, sweep, trim, rub\\
 & Levin ``dress verbs'' (original) & 0.027 & exercise, bathe, dress\\
 \hline
\multirow{5}{1em}{Domestic Animals} & Hypernym young.n.01 (original) & 0.033 & puppy, kitten, foal \\      
 & Hypernym domestic\_animal.n.01 (original) & 0.032 & puppy, retriever, pug\\
 & General Inquirer ``animal'' (replacement) & 0.023 & dog, animal, cat, goat \\
 & Hypernym canine.n.02 (replacement) & 0.021& puppy, retriever, pug\\
\midrule
\midrule
\multicolumn{4}{c}{\sc CLIP performs worse on} \\
\midrule
\midrule
\multirow{4}{1em}{Furniture} & Presence of word ``sofa'' (in common) & -0.032 & sofa\\
& Hypernym bedroom\_furniture.n.01 (in common) & -0.026 & bed, sofa\\
& Hypernym furniture.n.01 (in common) & -0.017 & couch, bed, sofa, chair, bench \\
& LIWC ``home'' (in common) & -0.015 & bed, window, sofa, room\\
\hline
\multirow{4}{1em}{Transportation} & Presence of word ``ride'' (original) & -0.027 & ride \\
& Hypernym vessel.n.02 (in common) & -0.019 & boat, ship, yacht\\
& Levin ``pedal'' verbs (original) & -0.018 & ride, drive, fly, sail, cruise\\
& Hypernym craft.n.02 (in common) & -0.018 & boat, balloon, ship, scooter, kayak\\ 
\hline
Herbivores & Hypernym ungulate.n.01 (in common) & -0.021 & horse, cow, camel, goat, deer\\
& Presence of word ``horse'' (in common) & -0.019 & horse\\
\hline
\multirow{3}{1em}{Sports} & Hypernym happening.n.01 (in common) & -0.021 & wave, win, tap, slam\\
& Hypernym contestant.n.01 (in common) & -0.020 & footballer, golfer, goalkeeper, cricketer, tackle\\
& Levin ``admire'' verbs (original) & -0.017 & stand, enjoy, admire, support\\
\hline
\multirow{2}{1em}{Academia} & General Inquirer ``academia'' (in common) & -0.020 & student, classroom, library, teacher, book, computer, conference \\
& Presence of word ``student'' (in common) & -0.020 & student \\
\bottomrule
\end{tabular}
\caption{CLIP relative performance analysis on a subset of binary features: the top-5 \textbf{easier} topics are \textit{Natural Phenomenon}, \textit{Waterfront Infrastructure},
\textit{Landscapes}, \textit{Grooming} and \textit{Domestic Animals}, while 
the top-5 \textbf{harder} topics are \textit{Furniture}, \textit{Transportation}, \textit{Herbivores}, \textit{Sports} and \textit{Academia}.}
\label{tab:categorical2}
\end{table*}

\paragraph{(8) CLIP performs relatively better on \textit{nature-related} and \textit{personal care} concepts and relatively worse on \textit{furniture, transportation, herbivores, sports, academia}.}
As mentioned in \cref{sec:model_output}, score \(D\) measures the relative CLIP performance, which is more relevant for retrieval models like CLIP.
Therefore, we measure the importance of each feature with respect to \(D\).
Specifically, we compute the mean differences of the \(D\) scores when the binary feature is present and when is not.
We show the CLIP performance analysis on \textbf{binary} features in \cref{tab:categorical2}.
Following the example of SEAL~\cite{Rajani2022SEALI}, we use ChatGPT to cluster the features under a broad topic automatically.\footnote{We use the following prompt: "Name a topic for the following words: \ldots{}"} 

We find that CLIP performs relatively \textbf{better} on topics related to nature: \textit{Natural Phenomenon},
\textit{Waterfront Infrastructure}, \textit{Landscapes}, \textit{Domestic Animals}, and personal care: \textit{Grooming},
and \textbf{worse} on topics like \textit{Furniture}, \textit{Transportation}, \textit{Herbivores}, \textit{Sports} and \textit{Academia}.

\section{Conclusion}

In this work, we proposed a simple and effective method to probe vision-language models. Our method is scalable, as it does not require data annotation and makes use of existing datasets. With our method, we analyzed the performance of CLIP, a popular state-of-the-art multi-modal model, on the SVO-Probes benchmark. 
We confirmed the recent findings of \citet{Thrush2022WinogroundPV} of CLIP behaving like a bag of words model and that of \citet{Parcalabescu2022VALSEAT} of CLIP performing better with nouns and verbs.
We also uncovered novel findings, for instance, that CLIP gets confused by concrete words, surprisingly improves performance for more ambiguous terms, or that the frequency of words does not significantly change the behavior of CLIP.

We hope our work contributes to ongoing efforts to discover the limitations of multi-modal models and help build more robust and reliable systems. 
Our framework can be easily used to analyze other benchmarks, features, and multi-modal models, and it is publicly available at
\url{https://github.com/MichiganNLP/Scalable-VLM-Probing}. 

\section*{Limitations}

SVO-Probes dataset is not balanced. For example, ``person'', ``man'', and ``woman'' are considerably more frequent than other words.
Future work can address this limitation by aggregating data from multiple datasets and balancing it out.
At the same time, the target dataset should reflect the phenomenon one wants to study.
For example, LAION~\cite{Schuhmann2021LAION400MOD} could be employed to study how VLMs perform with everyday human actions. Still, it may be too centered around objects (as opposed to actions) and overly noisy -- future work can consider using subsets instead.
A smaller yet cleaner alternative is Conceptual Captions~\cite{cc}.

Another limitation is not considering the polysemy when using LIWC or Levin dictionaries. This may lead to incorrect word categorization and influence the error analysis. Future work can mediate this limitation by linking semantic dictionaries such as Levin or LIWC with their WordNet synsets.

\section*{Acknowledgements}
We want to thank the anonymous reviewers for their helpful comments. We also thank Artem Abzaliev, Fabian Caba, Mohamed El Banani, and Karan Desai for the productive discussions.
This material is partly based on work supported by the Automotive Research Center (``ARC''). Any opinions, findings, conclusions, or recommendations expressed in this material are those of the authors and do not necessarily reflect the views of ARC or any other related entity.

\bibliography{main.bib}
\bibliographystyle{acl_natbib}

\end{document}